\documentclass{article}

\usepackage[preprint]{neurips_2019}

\usepackage[utf8]{inputenc} % allow utf-8 input
\usepackage[T1]{fontenc}    % use 8-bit T1 fonts
\usepackage{hyperref}       % hyperlinks
\usepackage{url}            % simple URL typesetting
\usepackage{booktabs}       % professional-quality tables
\usepackage{amsfonts}       % blackboard math symbols
\usepackage{nicefrac}       % compact symbols for 1/2, etc.
\usepackage{microtype}      % microtypography
\usepackage{subcaption}
\usepackage{graphicx}
\usepackage{amsmath}
\usepackage{comment}
\title{Continual Reinforcement Learning with Multi-Timescale Replay}

\author{
  Christos Kaplanis \\
  Department of Computing\\
  Imperial College London\\
  \texttt{christos.kaplanis14@imperial.ac.uk} \\
   \AND
  Claudia Clopath \thanks{Equal Contribution.} \\
  Department of Bioengineering \\
  Imperial College London \\
  \texttt{c.clopath@imperial.ac.uk} \\
    \And
  Murray Shanahan \textsuperscript{*} \\
  Department of Computing \\
  Imperial College London /\\
  DeepMind \\
  \texttt{m.shanahan@imperial.ac.uk} 
}

\begin{document}

\maketitle

\begin{abstract}
  In this paper, we propose a \textit{multi-timescale replay} (MTR) buffer for improving continual learning in RL agents faced with environments that are changing \textit{continuously} over time at timescales that are \textit{unknown} to the agent. The basic MTR buffer comprises a cascade of sub-buffers that accumulate experiences at different timescales, enabling the agent to improve the trade-off between adaptation to new data and retention of old knowledge. We also combine the MTR framework with invariant risk minimization \citep{arjovsky2019invariant} with the idea of encouraging the agent to learn a policy that is robust across the various environments it encounters over time. The MTR methods are evaluated in three different continual learning settings on two continuous control tasks and, in many cases, show improvement over the baselines.
\end{abstract}

\section{Introduction}

Artificially intelligent agents that are deployed in the real world have to be able to learn from data streaming in from a \textit{nonstationary} distribution and incrementally build on their knowledge over time, while operating with \textit{limited} computational resources; this is the challenge of \textit{continual learning} \citep{ring1994continual}. Artificial neural networks (ANNs), however, have long been known to suffer from the problem of \textit{catastrophic forgetting} \citep{mccloskey1989catastrophic}, whereby, in a context where the data distribution is changing over time, training on new data can result in abrupt erasure of previously acquired knowledge. This precludes them from being able to learn continually. %The rise of deep learning as a powerful machine learning technique has sparked a renewed urgency to solve the catastrophic forgetting problem and enable continual learning in neural-network based agents \citep{parisi2019continual}.

Typically, the testing of methods for mitigating catastrophic forgetting has been conducted in the context of training on a number of \textit{distinct} tasks in sequence. A consequence of this format for evaluation is that many continual learning techniques make use of the \textit{boundaries} between tasks in order to consolidate knowledge during training \citep{ruvolo2013ella, kirkpatrick2017overcoming, zenke2017continual}. In the real world, however, changes to the data distribution may happen \textit{gradually} and at \textit{unpredictable timescales}, in which case many of the existing techniques are simply not applicable, prompting the community to pose task-agnostic and task-free continual learning as desiderata for our agents \citep{clworkshop2016nips, clworkshop2018nips}. 

 \textit{Reinforcement learning} \citep{sutton1998reinforcement} is a paradigm that naturally poses these challenges, where the changes to the data distribution can occur unpredictably during the training of a \textit{single} task and can arise from multiple sources, e.g.: (i) correlations between successive states of the environment, (ii) changes to the agent's policy as it learns, and (iii) changes to the dynamics of the agent's environment. For these reasons, catastrophic forgetting can be an issue in the context in \textit{deep reinforcement learning}, where the agents are neural network-based. One commonly used method to tackle (i) is \textit{experience replay} (ER) \citep{lin1992self, mnih2015human}, whereby the agent's most recent experiences are stored in a first-in-first-out (FIFO) buffer, which is then sampled from \textit{at random} during training.  By shuffling the experiences in the buffer, the data are then \textit{identically} and \textit{independently distributed} (\textit{i.i.d.}) at training time, which prevents forgetting over the (short) timescale of the buffer since the distribution over this period is now \textit{stationary}.

\subsection{Experience replay for continual reinforcement learning}

The community has naturally investigated whether ER can be used to mitigate forgetting over the \textit{longer} timescales that are typically associated with continual learning, particularly because it does not necessarily require prior knowledge of the changes to the data distribution. One key observation that has been made in both a sequential multi-task setting \citep{isele2018selective, rolnick2019experience}, as well as in a single task setting \citep{de2015importance, de2016off, zhang2017deeper, wang2019boosting}, is that it is important to maintain a balance between the storage of \textit{new} and \textit{old} experiences in the buffer. By focusing just on recent experiences, the agent can easily forget what to do when it revisits states it has not seen in a while, resulting in catastrophic forgetting and instability; by retaining too many old experiences, on the other hand, the agent might focus too much on replaying states that are not relevant to its current policy, resulting in a sluggish and/or noisy improvement in its performance.

In this paper, we propose a \textit{multi-timescale replay} (MTR) buffer to improve continual reinforcement learning, which consists of a cascade of interacting sub-buffers, each of which accumulates experiences at a different timescale. It was designed with the following three motivating factors:
\begin{itemize}
    \item Several of the previously mentioned replay methods use just two timescales of memory in order to strike a balance between new and old experiences \citep{isele2018selective, rolnick2019experience, zhang2017deeper}. For example, one method in \citep{isele2018selective} combines a small FIFO buffer with a \textit{reservoir} buffer that maintains a uniform distribution over the agent's entire history of experiences \citep{vitter1985random} - this means that the composition of the replay database will adjust to short term changes in the distribution (with the FIFO buffer) and to long term changes (with the reservoir buffer), but it will not be as sensitive to medium term changes. Our method, on the other hand, maintains several sub-buffers that store experiences at a range of timescales, meaning that it can adjust well in scenarios where the rate of change of the distribution is unknown and can vary over time.
    \item The MTR buffer also draws inspiration from psychological evidence that the function relating the strength of a memory to its age follows a \textit{power law} \citep{wixted1991form}; forgetting tends to be fast soon after the memory is acquired, and then it proceeds slowly with a long tail. In the MTR buffer, as a result of the combination of multiple timescales of memory, the probability of a given experience lasting beyond a time $t$ in the database also follows a power law; in particular, it approximates a $\frac{1}{t}$ decay (Appendix \ref{app:powerlaw}).
    \item While shuffling the data to make it i.i.d. helps to prevent forgetting, it also discards structural information that may exist in the sequential progression of the data distribution - something that it is preserved to a degree in the cascade of sub-buffers in the MTR method. \textit{Invariant risk minimization} (IRM) is a recently developed method that uses the assumption that the training data has been split up into different \textit{environments} in order to train a model that is \textit{invariant} across these environments and is thus more likely to be robust and generalise well. In a second version of the MTR model, the \textit{MTR-IRM} agent, we apply the IRM principle by treating each sub-buffer as a different environment to see if it can improve continual learning by encouraging the agent to learn a policy that is invariant over time.
\end{itemize}

We test the two MTR methods in RL agents trained on continuous control tasks in a standard RL setting, as well as in settings where the environment dynamics are continuously modified over time. We find that the standard MTR agent is the best continual learner overall when compared to a number of baselines, and that the MTR-IRM agent improves continual learning in some of the more nonstationary settings, where one would expect an invariant policy to be more beneficial.

\section{Preliminaries}

\subsection{Soft Actor-Critic}

We used the soft actor-critic (SAC) algorithm \citep{haarnoja2018soft} for all experiments in this paper, adapting the version provided in \citep{stablebaselines}. SAC is an actor-critic RL algorithm based on the maximum entropy RL framework, which generalises the standard RL objective of maximising return by simultaneously maximising the entropy of the agent's policy:
\begin{equation}
\pi^* = \arg \max_{\pi} \mathbb{E}_{\pi} \left[ \sum_{t=0}^\infty  \gamma^t \left( r(s_t,a_t) + \alpha \mathcal{H}(\pi (\cdot | s_t)) \right)\right]
\end{equation}
where $s_t$ and $a_t$ represent the state visited and action taken at time $t$ respectively, $r(s_t, a_t)$ is the reward at time $t$, $\pi$ is a stochastic policy defined as a probability distribution over actions given state, $\pi^*$ represents the optimal policy, $\gamma$ is the discount factor, $\mathcal{H}$ is the entropy and $\alpha$ controls the balance between reward and entropy maximisation. This objective encourages the agent to find multiple ways of achieving its goal, resulting in more robust solutions. Robustness was a particularly important factor in choosing an RL algorithm for this project, since the added nonstationarity of the environment in two of the experimental settings can easily destabilise the agent's performance easily. In initial experiments, we found that SAC was more stable that other algorithms such as DDPG \citep{lillicrap2015continuous}. We used the automatic entropy regulariser used in \citep{haarnoja2018soft2}, which was found to be more robust than a fixed entropy regularisation coefficient.

\subsection{Invariant Risk Minimisation}

Invariant risk minimisation \citep{arjovsky2019invariant} is an approach that seeks to improve out-of-distribution generalisation in machine learning models by training them to learn \textit{invariant} or \textit{stable} predictors that avoid spurious correlations in the training data, a common problem with the framework of \textit{empirical} risk minimisation \citep{vapnik2006estimation}. While typically the training data and test data are randomly shuffled in order to ensure they are from the same distribution, IRM poses that information is actually \textit{lost} this way, and it starts with the assumption that the data can be split up into a number of different \textit{environments} $e \in \mathcal{E}_{tr}$. The IRM loss function encourages the model to learn a mapping that is invariant across all the different training environments, with the hope that, if it is stable across these, then it is more likely to perform well in previously unseen environments. The IRM loss is constructed as follows:
\begin{equation}
\min_{\Phi: \mathcal{X}\rightarrow \mathcal{Y}} \sum_{e \in \mathcal{E}_{tr}} R^e (\Phi) + \lambda \cdot || \nabla_{w | w=1.0} R^e (w \cdot \Phi) ||^2
\end{equation}
where $\Phi$ is the mapping induced by the model that maps the inputs to the outputs (and is a function of the model parameters), $R^e$ is the loss function for environment $e$, $w$ is a dummy variable and $\lambda$ is a parameter that balances the importance of the empirical loss (the first term) and the IRM loss (the second term). The goal of the IRM loss is to find a representation $\Phi$ such that the optimal readout $w$ is the same (i.e. the gradient of the readout is zero), no matter the environment; this way, when a new environment is encountered, it is less likely that the policy will have to \textit{change} in order to suit it. In the next section, we describe how the IRM principle is applied in one version of the MTR replay database where the different environments correspond to experiences collected at different timescales.

\section{Multi-Timescale Replay}

\begin{figure}[h]
\centering
\begin{subfigure}{0.6 \textwidth}
\includegraphics[width=\textwidth]{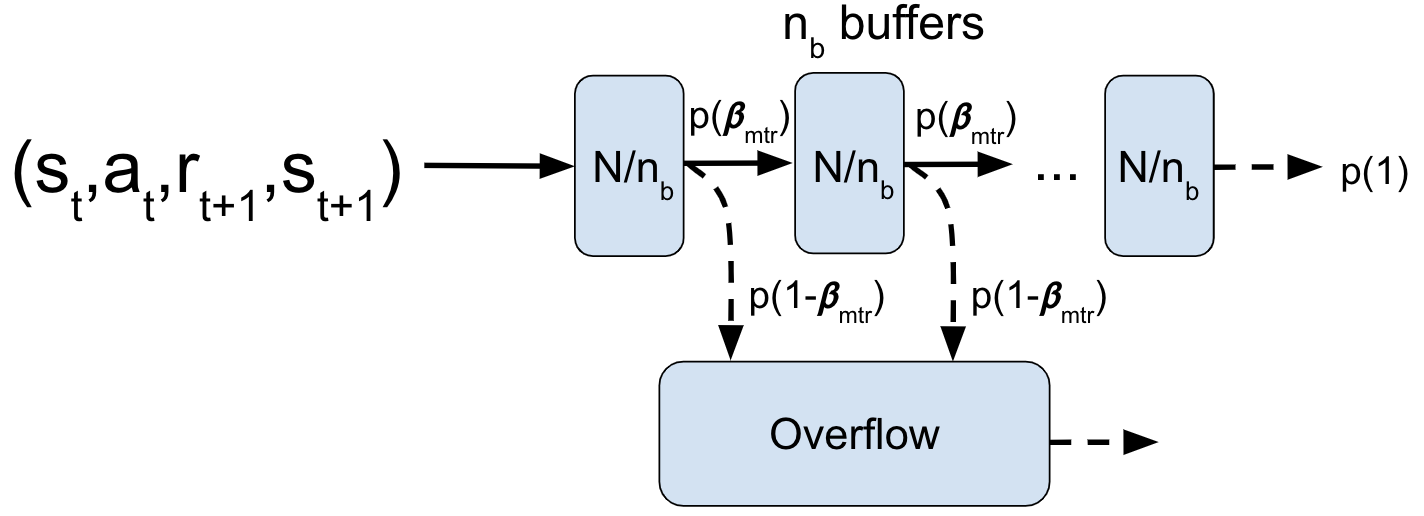}
\caption{}
\end{subfigure}
\begin{subfigure}{0.39 \textwidth}
\includegraphics[width=0.8\textwidth]{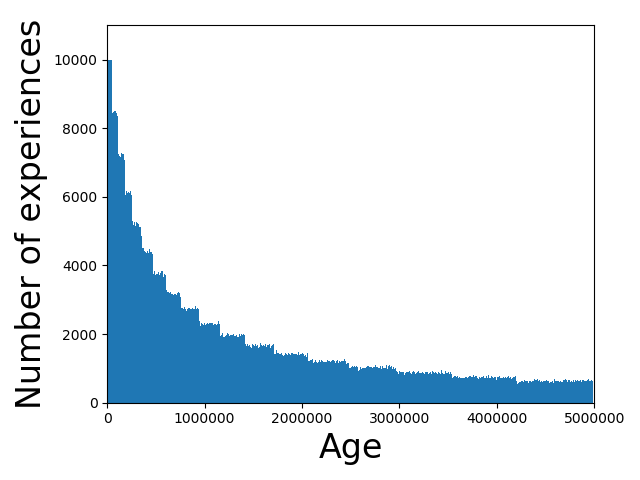}
\caption{}
\end{subfigure}
\caption{(a) Diagram of MTR buffer; each blue box corresponds to a FIFO sub-buffer. (b) Histogram of age of experiences in cascade of MTR buffer after 5 million experiences inserted.}
\label{fig:mtr_diagram}
\end{figure}

The multi-timescale replay (MTR) database of size $N$ consists of a cascade of $n_b$ FIFO buffers, each with maximum size $\frac{N}{n_b}$, and a separate overflow buffer (also FIFO), which has a dynamic maximum size that is equal to the difference between $N$ and the number of experiences currently stored in the cascade (Figure \ref{fig:mtr_diagram}(a)). New experiences of the form $(s_t, a_t, r_{t+1}, s_{t+1})$ are pushed into the first sub-buffer. When a given sub-buffer is full, the oldest experience in the buffer is pushed out, at which point it has two possible fates: (i) with a predefined probability $\beta_{mtr}$ it gets pushed into the next sub-buffer, or (ii) with probability $1-\beta_{mtr}$ it gets added to the overflow buffer. If the total number of experiences stored in the cascade and the overflow buffer exceeds $N$, the overflow buffer is shrunk with the oldest experiences being removed until the database has at most $N$ experiences. Once the cascade of sub-buffers is full, the size of the overflow buffer will be zero and any experience that is pushed out of any of the sub-buffers is discarded. During training, the number of experiences sampled from each sub-buffer (including the overflow buffer) is proportional to the fraction of the total number of experiences in the database contained in the sub-buffer. Figure \ref{fig:mtr_diagram}(b) shows the distribution of ages of experiences in the MTR buffer, and a mathematical intuition for how the MTR buffer results in a power law distribution of memories is given in Appendix \ref{app:powerlaw}.

\subsection{IRM version of MTR}

In the IRM version of the MTR agent (MTR-IRM), we assume that the set of experiences in each sub-buffer of the MTR cascade corresponds to data collected in a different environment. Under this assumption, we can apply the IRM principle to the \textit{policy network} of the SAC agent, so each $R^e (\Phi)$ corresponds to the policy loss calculated using the data in the corresponding sub-buffer. While it would be interesting to apply the IRM to the value losses too, in this work, for simplicity we only applied it to the policy loss of the agent. The per-experience policy loss for SAC is as follows:
\begin{equation}
L_{\pi}(\phi, s) = \mathbb{E}_{\mathbf{a} \sim \pi_\phi} \left[ \alpha_t \log ( \pi_\phi (a | s)) - Q_{\theta_1}(s,a) \right]
\end{equation}
where $\phi$ are the parameters of the policy network, $\pi_\phi$ is the conditional action distribution implied by the policy network, $s$ is the state, $a$ is the action sampled from $\pi_\phi$, $\alpha_t$ is the dynamic entropy coefficient and $Q_{\theta_1}$ is the Q-value function implied by one of the two Q-value networks used in SAC. The policy loss at each iteration is calculated by averaging the per-experience loss shown above over a mini-batch of experiences chosen from the replay database. In combination with IRM, however, the overall policy loss is as evaluated as follows:
\begin{equation}
L_{\pi_{IRM}}(\phi) = \mathbb{E}_{s_t \sim \mathcal{D}}\left[L_{\pi}(\phi, s_t)\right] + \lambda_{IRM} \cdot \sum_{i=1}^{n_b} \frac{| \mathcal{D}_i |}{| \mathcal{D}_{cascade}|} \mathbb{E}_{s_{t'} \sim \mathcal{D}_i}\left[ || \nabla_{w | w=1.0} L_{\pi}(\phi, s_{t'}, w) || ^2 \right]
\end{equation}
where $|\mathcal{D}_i|$ is the number of experiences in the $i^{th}$ sub-buffer in the cascade, $|\mathcal{D}_{cascade}|$ is the total number of experiences stored in the cascade of sub-buffers, $w$ is a dummy variable, and $L_{\pi}$ is overloaded, such that:
\begin{equation}
L_{\pi}(\phi,s_{t'}, w) = \mathbb{E}_{a \sim \pi_\phi} \left[ \alpha_t \log ( \pi_\phi (w \cdot a | s)) - Q_{\theta_1}(s, a) \right]
\end{equation}
The balance between the original policy loss and the extra IRM constraint is controlled by $\lambda_{IRM}$. 

\section{Experiments}

\subsection{Setup}
 
The two MTR methods were evaluated in RL agents trained with SAC \citep{haarnoja2018soft} on two different continuous control tasks (RoboschoolAnt and RoboschoolHalfCheetah)\footnote{These environments are originally from \citep{schulman2017proximal} and were modified by adapting code from \citep{packer2018assessing}.}, where the strength of \textit{gravity} in the environments was modified continuously throughout training in three different ways: fixed gravity, linearly increasing gravity, and gravity fluctuating in a sine wave. The idea was to see how the agents could cope with changes to the distribution that arise from different sources and at different timescales. In the fixed gravity setting, which constitutes the standard RL setup, the changes to the distribution result from correlation between successive states and changes to the agent's policy. In the other two settings, continual learning is made more challenging because changes are also made to the dynamics of the environment. In the linear setting, the gravity is adjusted slowly over time, with no setting being repeated at any point during training; in the fluctuating setting, the changes are faster and gravity settings are revisited so that the \textit{relearning} ability of the agent can be observed.

In order to evaluate the continual learning ability of the agents, their performance was recorded over the course of training (in terms of mean reward) on (i) the current task at hand, and (ii) on an evenly spaced subset of the gravity environments experienced during training ($-7$, $-9.5$, $-12$, $-14.5$ and $-17 m/s^2$). It is worth noting that initial experiments were run in a traditional \textit{multi-task} setting, where the gravity was uniformly sampled from an interval of $[-17, -7]$ throughout training, i.e. the tasks are \textit{interleaved}, in order to ensure that it is possible to learn good policies for all tasks in the same policy network (Figure \ref{fig:mtr_multitask}). Further experimental details and a table of the hyperparameters used for training are given in Appendix \ref{app:details}.

%In the linearly increasing and fluctuating gravity settings the gravity is modified with in a range of -7 to -17 ms^-2

\subsection{Results}

Across all three continual learning settings, the basic MTR agent appears to be the most consistent performer, demonstrating either the best or second best results in terms of training reward and mean evaluation reward in all tasks, indicating that recording experiences over multiple timescales can improve the tradeoff between new learning and retention of old knowledge in RL agents. The MTR-IRM agent achieved the best evaluation reward in two of the more nonstationary settings for the HalfCheetah task, but not in the Ant task, indicating that learning a policy that is invariant across time can be beneficial for generalisation and mitigating forgetting, but that it might depend on the particular task setting and the transfer potential between the different environments. Below, we discuss the results from each setting in more detail. All plots show moving averages of mean reward over three runs per agent type with standard error bars.

\begin{figure}[h]
\centering
\begin{subfigure}{0.49\textwidth}
\includegraphics[width=\textwidth]{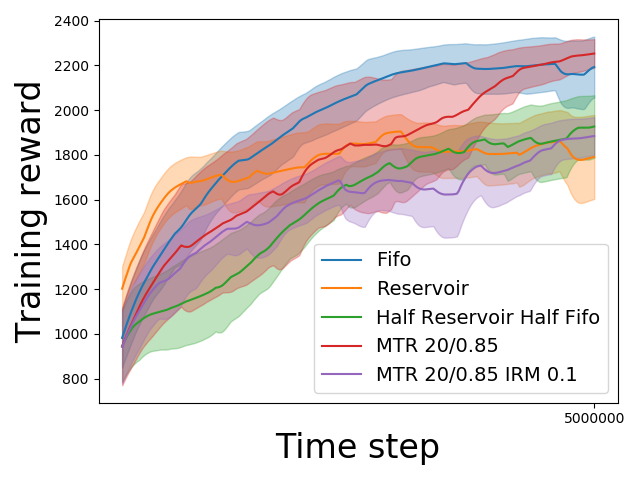}
\caption{HalfCheetah Train}
\end{subfigure}
\begin{subfigure}{0.49\textwidth}
\includegraphics[width=\textwidth]{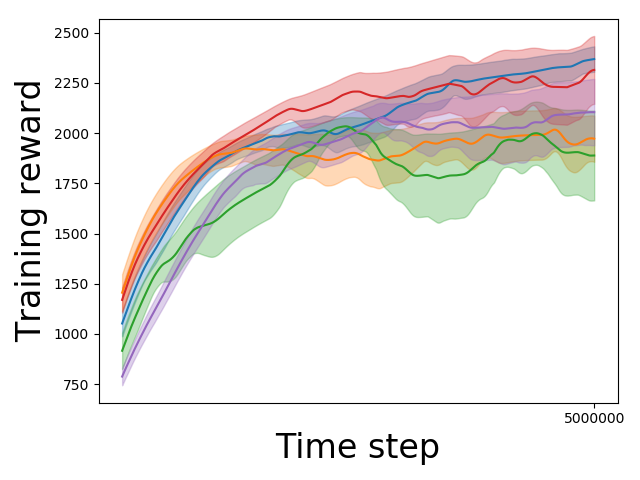}
\caption{Ant Train}
\end{subfigure}
\caption{Fixed gravity setting ($-9.81m/s^2$). Training reward for (a) HalfCheetah and (b) Ant.}
\label{fig:fixed_gravity_results}
\end{figure}

\paragraph{Fixed gravity} In the fixed gravity experiments, the FIFO and MTR agents were consistently the best performers (Figure \ref{fig:fixed_gravity_results}), with both agents achieving a similar final level of training reward in both the HalfCheetah and Ant tasks. One would expect the FIFO agent to be a relatively strong performer in this setting, since the environment dynamics are stable and so the retention of old memories is likely to be less crucial than in the other two more nonstationary settings. The fact that the basic MTR agent performs as well as the FIFO agent shows that the replay of some older experiences is not holding back the progress of the agent, but also that it does not seem to particularly help the overall performance either. The MTR-IRM agent, on the other hand, performed poorly in the fixed gravity setting, presumably because there is not enough nonstationarity to reap any generalisation benefits from learning an invariant representation for the policy, and instead the IRM constraints just slow down the pace of improvement of the agent.

\paragraph{Linearly increasing gravity} In the linearly increasing gravity experiments, the FIFO agent performed best in terms of training reward, but was the \textit{worst} performer when evaluated on the 5 different gravity settings on both tasks. This is somewhat intuitive: the FIFO agent can be expected to do well on the training task as it is only training with the most recent data, which are the most pertinent to the task at hand; on the other hand, it quickly forgets what to do in gravity settings that it experienced a long time ago (Figure \ref{fig:individual_eval_results}(a)). In the HalfCheetah task, the MTR-IRM agent surpassed all other agents in the evaluation setting by the end of training, with the standard MTR agent coming second, perhaps indicating that, in a more nonstationary setting (in contrast with the fixed gravity experiments), learning a policy that is invariant over time can lead to a better overall performance in different environments. It was difficult, however, to identify any presence of \textit{forward transfer} in the MTR-IRM agent in plotting the individual evaluations rewards over time (Appendix \ref{app:additional}).

\begin{figure}[h]
\centering
\begin{subfigure}{0.49\textwidth}
\includegraphics[width=\textwidth]{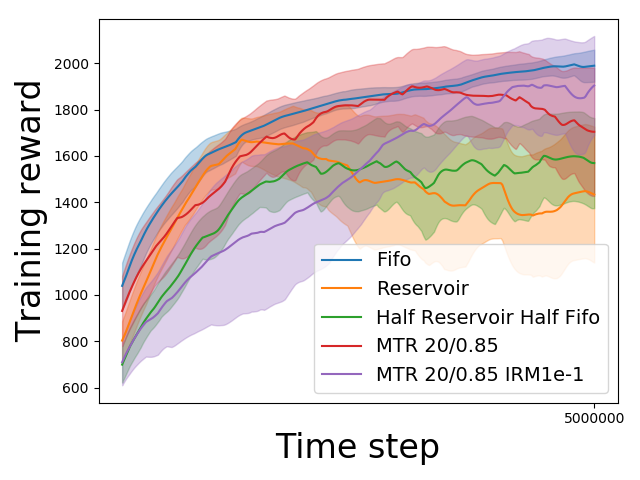}
\caption{HalfCheetah Train}
\end{subfigure}
\begin{subfigure}{0.49\textwidth}
\includegraphics[width=\textwidth]{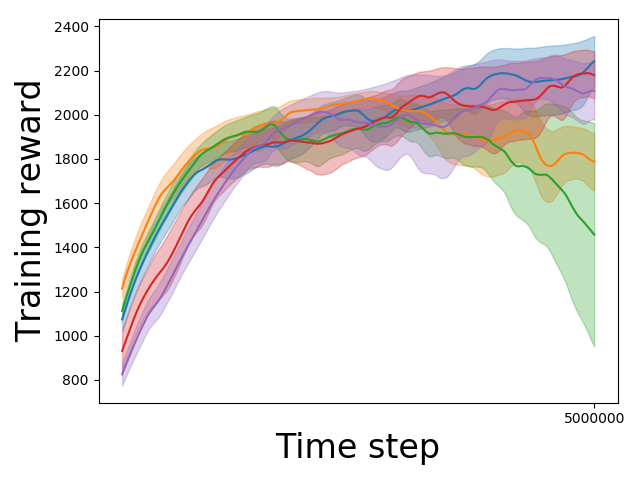}
\caption{Ant Train}
\end{subfigure}
\begin{subfigure}{0.49\textwidth}
\includegraphics[width=\textwidth]{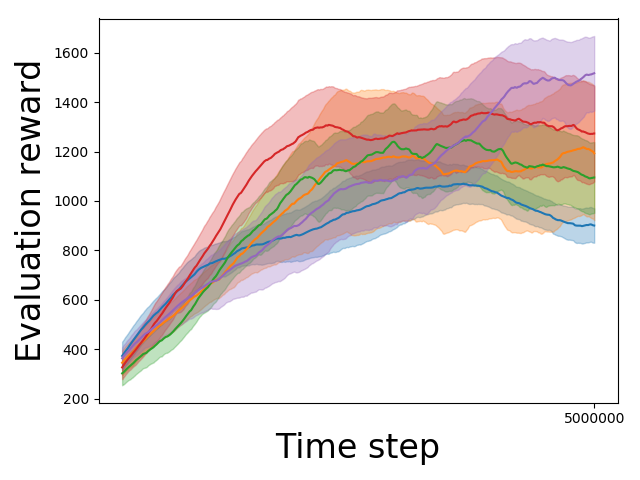}
\caption{HalfCheetah Mean Eval}
\end{subfigure}
\begin{subfigure}{0.49\textwidth}
\includegraphics[width=\textwidth]{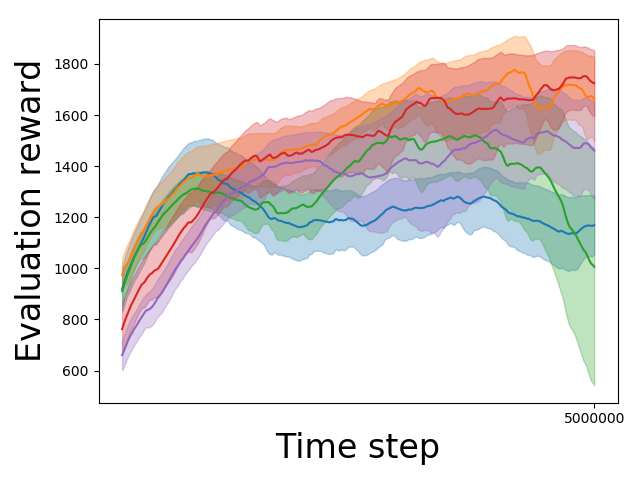}
\caption{Ant Mean Eval}
\end{subfigure}
\caption{Linearly increasing gravity setting. (Top) Training reward for (a) HalfCheetah and (b) Ant. (Bottom) Mean evaluation reward for (c) HalfCheetah and (d) Ant.}
\label{fig:linear_gravity_results}
\end{figure}

In the linear gravity setting for the Ant task, the FIFO, MTR and MTR-IRM agents were equally strong on the training task (Figure \ref{fig:linear_gravity_results}(b)), and the MTR and reservoir agents were joint best with regards to the mean evaluation reward (Figure \ref{fig:linear_gravity_results}(d)). The MTR-IRM agent does not show the same benefits as in the HalfCheetah setting; this could be because it is more difficult to learn an invariant policy across the different gravity settings across this task, with less potential for transfer between policies for the various environments. The transferability of policies across different environments and the effects of the order in which they are encountered are important aspects for future investigation.

\begin{figure}[h]
\centering
\begin{subfigure}{0.49\textwidth}
\includegraphics[width=\textwidth]{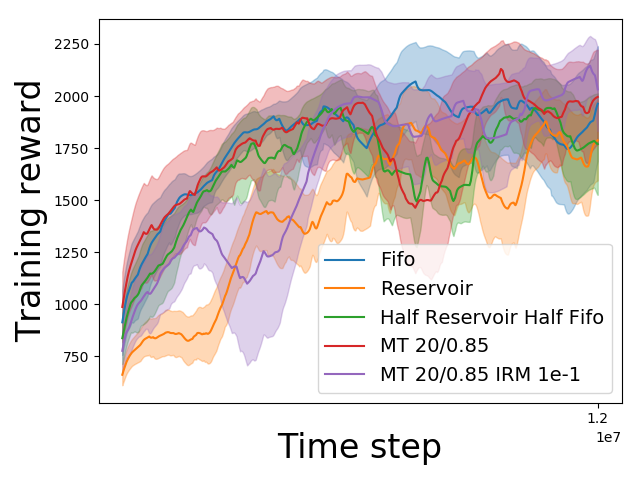}
\caption{HalfCheetah Train}
\end{subfigure}
\begin{subfigure}{0.49\textwidth}
\includegraphics[width=\textwidth]{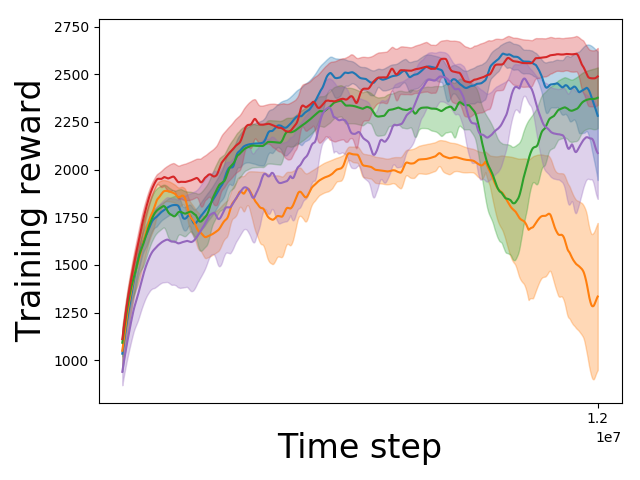}
\caption{Ant Train}
\end{subfigure}
\begin{subfigure}{0.49\textwidth}
\includegraphics[width=\textwidth]{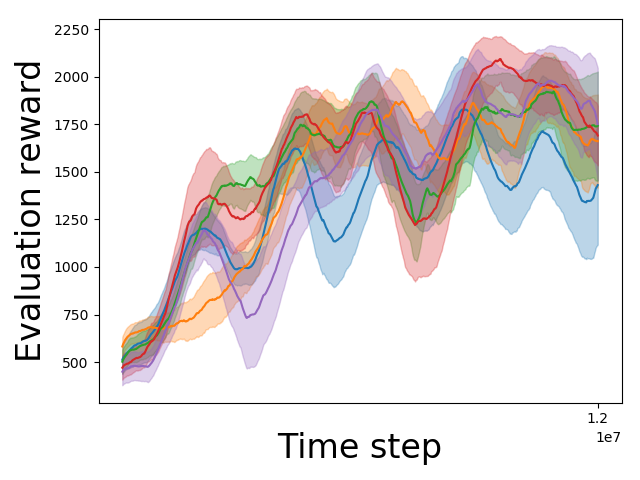}
\caption{HalfCheetah Mean Eval}
\end{subfigure}
\begin{subfigure}{0.49\textwidth}
\includegraphics[width=\textwidth]{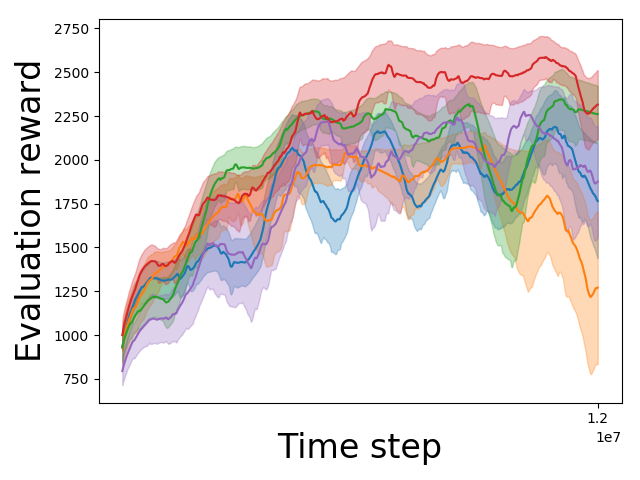}
\caption{Ant Mean Eval}
\end{subfigure}
\caption{Fluctuating gravity setting. (Top) Training reward for (a) HalfCheetah and (b) Ant. (Bottom) Mean evaluation reward for (c) HalfCheetah and (d) Ant.}
\label{fig:fluctuating_gravity_results}
\end{figure}

\paragraph{Fluctuating gravity} In the fluctuating gravity setting, the performances of the various agents were less differentiated than in the linearly increasing gravity setting, perhaps because the timescale of changes to the distribution were \textit{shorter} and the agents had the opportunity to \textit{revisit} gravity environments (Figure \ref{fig:fluctuating_gravity_results}). In the HalfCheetah task, the MTR-IRM agent was the best performer in terms of final training and evaluation rewards, though by a very small margin. In the Ant task, the best performer was the standard MTR agent, which reached a higher and more stable evaluation reward than any of the other agents. Once again, as in the linearly increasing gravity setting, the MTR-IRM agent struggled comparatively on the Ant task.

\begin{figure}[h]
\centering
\begin{subfigure}{0.49\textwidth}
\includegraphics[width=\textwidth]{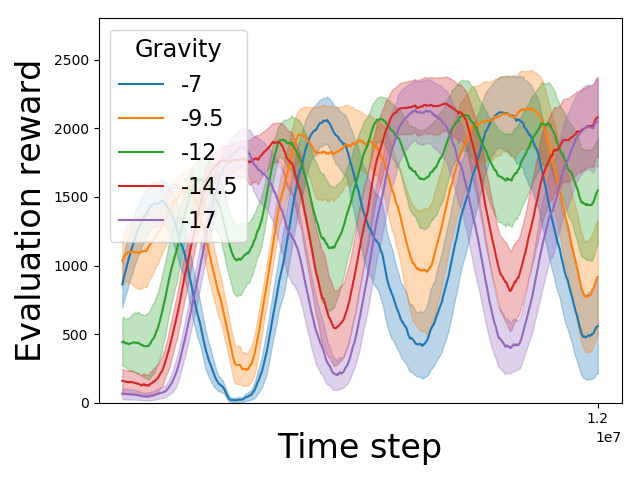}
\caption{FIFO}
\end{subfigure}
\begin{subfigure}{0.49\textwidth}
\includegraphics[width=\textwidth]{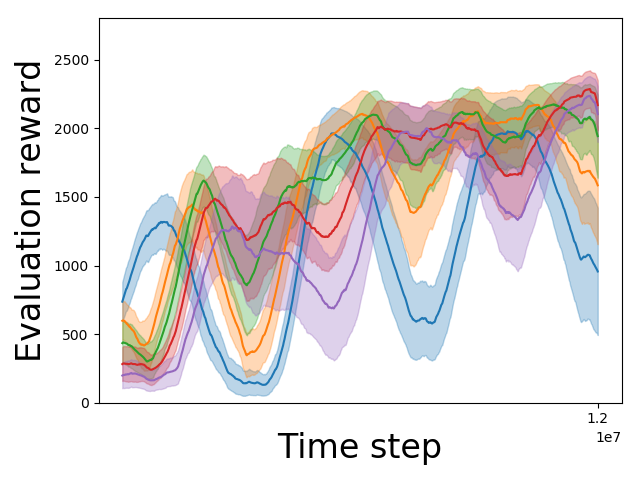}
\caption{MTR-IRM}
\end{subfigure}
\caption{Individual Evaluation rewards for fluctuating gravity HalfCheetah with (a) FIFO buffer and (b) MTR-IRM buffer.}
\label{fig:individual_eval_fluctuating}
\end{figure}

An interesting observation with regards to the agents' ability to relearn can be made by comparing the individual evaluation rewards of the FIFO and MTR-IRM agents in the fluctuating gravity setting. The fluctuating performance on each of the different gravity evaluation settings can be observed very clearly in the results of the FIFO agent (Figure \ref{fig:individual_eval_fluctuating}(a)), where the ups and downs in performance reflect the fluctuations of the gravity setting being trained on. While in the MTR-IRM agent, these fluctuations in performance can also be observed, the \textit{dips} in performance on gravity settings that have not been experienced in a while become significantly \textit{shallower} as training progresses, providing evidence that the agent is consolidating its knowledge over time (Figure \ref{fig:individual_eval_fluctuating}(b)).

\section{Related Work}

Many existing approaches for mitigating catastrophic forgetting in neural networks use buffers for storing past data or experiences, and are collectively known as \textit{replay-based methods} \citep{robins1995catastrophic, lopez2017gradient, chaudhry2019continual}. Here we briefly elaborate on a selection of works that investigate or make use of multiple timescales in the replay database, either in continual learning or in the standard RL setting. In \citep{de2015importance}, it is shown that retaining older experiences as well as the most recent ones can improve the performance of deep RL agents on the pendulum swing-up task, particularly for smaller replay databases. In \citep{zhang2017deeper}, it is shown that \textit{combined experience replay}, which trains agents on a combination of the most recent experiences as they come in and those in the replay buffer, enables faster learning than training on just one or the other, particularly when the replay database is very large. In \citep{isele2018selective}, various experience selection methods are evaluated for deep RL, and it was noted that for each method, a small FIFO buffer was maintained in parallel in order to ensure that the agent did not overfit to the more long-term memory buffer and had a chance to train on all experiences. In \citep{rolnick2019experience}, it is shown that a 50/50 split of fresh experiences and experiences sampled from a reservoir buffer provides a good balance between mitigating forgetting and reaching a high overall level of performance on different tasks in a deep RL setting. As discussed in the Introduction, these methods employ two different timescales in the distribution of experiences used for training, while the MTR methods use multiple timescales, which makes them sensitive to changes to the data distribution that occur at a range of different speeds or frequencies. 

In \citep{wang2019boosting}, it is shown that prioritising the replay of recent experiences in the buffer improves the performance of deep RL agents, using a \textit{continuum} of priorities across time. In this paper, a FIFO buffer is used, so the data is only \textit{stored} at the most recent timescale, but the probability of an experience being chosen for replay decays exponentially with its age. Other non-replay-based approaches that use multiple timescales of memory to mitigate catastrophic forgetting in deep RL include \citep{kaplanis2018continual}, which consolidates the individual parameters at multiple timescales, and \citep{kaplanis2019policy}, which consolidates the agent's policy at multiple timescales, using a cascade of hidden policy networks.

\section{Conclusion}

In this paper, we investigated whether a replay buffer set up to record experiences at multiple timescales could help in a continual reinforcement learning setting where the timing, timescale and nature of changes to the incoming data distribution are unknown to the agent. One of the versions of the multi-timescale replay database was combined with the invariant risk minimisation principle \citep{arjovsky2019invariant} in order to try and learn a policy that is \textit{invariant} across time, with the idea that it might lead to a more robust policy that is more resistant to catastrophic forgetting. We tested numerous agents on two different continuous control tasks in three different continual learning settings and found that the basic MTR agent was the most consistent performer overall. The MTR-IRM agent was the best continual learner in two of the more nonstationary settings in the HalfCheetah task, but was relatively poor on the Ant task, indicating that the utility of the IRM principle may depend on specific aspects of the tasks at hand and the transferability between the policies in different environments.

\paragraph{Future Work} One important avenue for future work would be to evaluate the MTR model in a broader range of training settings, for example to vary the timescales at which the environment is adjusted (e.g. gravity fluctuations at different frequencies). Furthermore, it would be useful to evaluate the sensitivity of the MTR method's performance to its hyperparameters ($\beta_{mtr}$ and $n_b$). Finally, it is worth noting that, in its current incarnation, the MTR method does not select which memories to retain for longer in an intelligent way - it is simply determined with a coin toss. In this light, it would be interesting to explore ways of \textit{prioritising} the retention of certain memories from one sub-buffer to the next, for example by the temporal difference error, which is used in \citep{schaul2015prioritized} to prioritise the \textit{replay} of memories in the buffer. 

\section{Code}

The code for this project is available at \url{https://github.com/ChristosKap/multi_timescale_replay}.

\bibliography{neurips_2019}

\begin{thebibliography}{34}
\providecommand{\natexlab}[1]{#1}
\providecommand{\url}[1]{\texttt{#1}}
\expandafter\ifx\csname urlstyle\endcsname\relax
  \providecommand{\doi}[1]{doi: #1}\else
  \providecommand{\doi}{doi: \begingroup \urlstyle{rm}\Url}\fi

\bibitem[clw(2016)]{clworkshop2016nips}
{Continual Learning} and {D}eep {N}etworks {W}orkshop.
\newblock \url{https://sites.google.com/site/cldlnips2016/home}, 2016.

\bibitem[clw(2018)]{clworkshop2018nips}
{Continual Learning} {W}orkshop.
\newblock \url{https://sites.google.com/view/continual2018/home}, 2018.

\bibitem[Arjovsky et~al.(2019)Arjovsky, Bottou, Gulrajani, and
  Lopez-Paz]{arjovsky2019invariant}
Martin Arjovsky, L{\'e}on Bottou, Ishaan Gulrajani, and David Lopez-Paz.
\newblock Invariant risk minimization.
\newblock \emph{arXiv preprint arXiv:1907.02893}, 2019.

\bibitem[Chaudhry et~al.(2019)Chaudhry, Rohrbach, Elhoseiny, Ajanthan, Dokania,
  Torr, and Ranzato]{chaudhry2019continual}
Arslan Chaudhry, Marcus Rohrbach, Mohamed Elhoseiny, Thalaiyasingam Ajanthan,
  Puneet~K Dokania, Philip~HS Torr, and Marc'Aurelio Ranzato.
\newblock Continual learning with tiny episodic memories.
\newblock \emph{arXiv preprint arXiv:1902.10486}, 2019.

\bibitem[de~Bruin et~al.(2015)de~Bruin, Kober, Tuyls, and
  Babu{\v{s}}ka]{de2015importance}
T.~de~Bruin, J.~Kober, K.~Tuyls, and R.~Babu{\v{s}}ka.
\newblock The importance of experience replay database composition in deep
  reinforcement learning.
\newblock In \emph{Deep Reinforcement Learning Workshop, Advances in Neural
  Information Processing Systems (NIPS)}, 2015.

\bibitem[de~Bruin et~al.(2016)de~Bruin, Kober, Tuyls, and
  Babu{\v{s}}ka]{de2016off}
T.~de~Bruin, J.~Kober, K.~Tuyls, and R.~Babu{\v{s}}ka.
\newblock Off-policy experience retention for deep actor-critic learning.
\newblock In \emph{Deep Reinforcement Learning Workshop, Advances in Neural
  Information Processing Systems (NIPS)}, 2016.

\bibitem[Haarnoja et~al.(2018{\natexlab{a}})Haarnoja, Zhou, Abbeel, and
  Levine]{haarnoja2018soft}
Tuomas Haarnoja, Aurick Zhou, Pieter Abbeel, and Sergey Levine.
\newblock Soft actor-critic: Off-policy maximum entropy deep reinforcement
  learning with a stochastic actor.
\newblock In \emph{International Conference on Machine Learning}, pages
  1861--1870, 2018{\natexlab{a}}.

\bibitem[Haarnoja et~al.(2018{\natexlab{b}})Haarnoja, Zhou, Hartikainen,
  Tucker, Ha, Tan, Kumar, Zhu, Gupta, Abbeel, et~al.]{haarnoja2018soft2}
Tuomas Haarnoja, Aurick Zhou, Kristian Hartikainen, George Tucker, Sehoon Ha,
  Jie Tan, Vikash Kumar, Henry Zhu, Abhishek Gupta, Pieter Abbeel, et~al.
\newblock Soft actor-critic algorithms and applications.
\newblock \emph{arXiv preprint arXiv:1812.05905}, 2018{\natexlab{b}}.

\bibitem[Hill et~al.(2018)Hill, Raffin, Ernestus, Gleave, Kanervisto, Traore,
  Dhariwal, Hesse, Klimov, Nichol, Plappert, Radford, Schulman, Sidor, and
  Wu]{stablebaselines}
Ashley Hill, Antonin Raffin, Maximilian Ernestus, Adam Gleave, Anssi
  Kanervisto, Rene Traore, Prafulla Dhariwal, Christopher Hesse, Oleg Klimov,
  Alex Nichol, Matthias Plappert, Alec Radford, John Schulman, Szymon Sidor,
  and Yuhuai Wu.
\newblock Stable baselines.
\newblock \url{https://github.com/hill-a/stable-baselines}, 2018.

\bibitem[Isele and Cosgun(2018)]{isele2018selective}
David Isele and Akansel Cosgun.
\newblock Selective experience replay for lifelong learning.
\newblock In \emph{Thirty-Second AAAI Conference on Artificial Intelligence},
  2018.

\bibitem[Kahana and Adler(2017)]{kahana2017note}
Michael~J Kahana and Mark Adler.
\newblock Note on the power law of forgetting.
\newblock \emph{bioRxiv}, page 173765, 2017.

\bibitem[Kaplanis et~al.(2018)Kaplanis, Shanahan, and
  Clopath]{kaplanis2018continual}
Christos Kaplanis, Murray Shanahan, and Claudia Clopath.
\newblock Continual reinforcement learning with complex synapses.
\newblock In Jennifer Dy and Andreas Krause, editors, \emph{Proceedings of the
  35th International Conference on Machine Learning}, volume~80 of
  \emph{Proceedings of Machine Learning Research}, pages 2497--2506,
  Stockholmsmässan, Stockholm Sweden, 10--15 Jul 2018. PMLR.
\newblock URL \url{http://proceedings.mlr.press/v80/kaplanis18a.html}.

\bibitem[Kaplanis et~al.(2019)Kaplanis, Shanahan, and
  Clopath]{kaplanis2019policy}
Christos Kaplanis, Murray Shanahan, and Claudia Clopath.
\newblock Policy consolidation for continual reinforcement learning.
\newblock In Kamalika Chaudhuri and Ruslan Salakhutdinov, editors,
  \emph{Proceedings of the 36th International Conference on Machine Learning},
  volume~97 of \emph{Proceedings of Machine Learning Research}, pages
  3242--3251, Long Beach, California, USA, 09--15 Jun 2019. PMLR.
\newblock URL \url{http://proceedings.mlr.press/v97/kaplanis19a.html}.

\bibitem[Kirkpatrick et~al.(2017)Kirkpatrick, Pascanu, Rabinowitz, Veness,
  Desjardins, Rusu, Milan, Quan, Ramalho, Grabska-Barwinska,
  et~al.]{kirkpatrick2017overcoming}
J.~Kirkpatrick, R.~Pascanu, N.~Rabinowitz, J.~Veness, G.~Desjardins, A.~A.
  Rusu, K.~Milan, J.~Quan, T.~Ramalho, A.~Grabska-Barwinska, et~al.
\newblock Overcoming catastrophic forgetting in neural networks.
\newblock \emph{Proceedings of the National Academy of Sciences}, 114\penalty0
  (13):\penalty0 3521--3526, 2017.

\bibitem[Lillicrap et~al.(2015)Lillicrap, Hunt, Pritzel, Heess, Erez, Tassa,
  Silver, and Wierstra]{lillicrap2015continuous}
Timothy~P Lillicrap, Jonathan~J Hunt, Alexander Pritzel, Nicolas Heess, Tom
  Erez, Yuval Tassa, David Silver, and Daan Wierstra.
\newblock Continuous control with deep reinforcement learning.
\newblock \emph{arXiv preprint arXiv:1509.02971}, 2015.

\bibitem[Lin(1992)]{lin1992self}
Long-Ji Lin.
\newblock Self-improving reactive agents based on reinforcement learning,
  planning and teaching.
\newblock \emph{Machine learning}, 8\penalty0 (3-4):\penalty0 293--321, 1992.

\bibitem[Lopez-Paz et~al.(2017)]{lopez2017gradient}
David Lopez-Paz et~al.
\newblock Gradient episodic memory for continual learning.
\newblock In \emph{Advances in Neural Information Processing Systems}, pages
  6467--6476, 2017.

\bibitem[McCloskey and Cohen(1989)]{mccloskey1989catastrophic}
M.~McCloskey and J.~N. Cohen.
\newblock Catastrophic interference in connectionist networks: The sequential
  learning problem.
\newblock \emph{Psychology of Learning and Motivation}, 24:\penalty0 109--165,
  1989.

\bibitem[Mnih et~al.(2015)Mnih, Kavukcuoglu, Silver, Rusu, Veness, Bellemare,
  Graves, Riedmiller, Fidjeland, Ostrovski, et~al.]{mnih2015human}
V.~Mnih, K.~Kavukcuoglu, D.~Silver, A.~A. Rusu, J.~Veness, M.~G. Bellemare,
  A.~Graves, M.~Riedmiller, A.~K. Fidjeland, G.~Ostrovski, et~al.
\newblock Human-level control through deep reinforcement learning.
\newblock \emph{Nature}, 518\penalty0 (7540):\penalty0 529--533, 2015.

\bibitem[Packer et~al.(2018)Packer, Gao, Kos, Kr\"ahenb\"uhl, Koltun, and
  Song]{packer2018assessing}
Charles Packer, Katelyn Gao, Jernej Kos, Philipp Kr\"ahenb\"uhl, Vladlen
  Koltun, and Dawn Song.
\newblock Assessing generalization in deep reinforcement learning, 2018.

\bibitem[Ring(1994)]{ring1994continual}
Mark~Bishop Ring.
\newblock \emph{Continual learning in reinforcement environments}.
\newblock PhD thesis, University of Texas at Austin Austin, Texas 78712, 1994.

\bibitem[Robins(1995)]{robins1995catastrophic}
Anthony Robins.
\newblock Catastrophic forgetting, rehearsal and pseudorehearsal.
\newblock \emph{Connection Science}, 7\penalty0 (2):\penalty0 123--146, 1995.

\bibitem[Rolnick et~al.(2019)Rolnick, Ahuja, Schwarz, Lillicrap, and
  Wayne]{rolnick2019experience}
David Rolnick, Arun Ahuja, Jonathan Schwarz, Timothy Lillicrap, and Gregory
  Wayne.
\newblock Experience replay for continual learning.
\newblock In \emph{Advances in Neural Information Processing Systems}, pages
  348--358, 2019.

\bibitem[Rubin and Wenzel(1996)]{rubin1996one}
David~C Rubin and Amy~E Wenzel.
\newblock One hundred years of forgetting: A quantitative description of
  retention.
\newblock \emph{Psychological review}, 103\penalty0 (4):\penalty0 734, 1996.

\bibitem[Ruvolo and Eaton(2013)]{ruvolo2013ella}
P.~Ruvolo and E.~Eaton.
\newblock Ella: An efficient lifelong learning algorithm.
\newblock In \emph{International Conference on Machine Learning}, pages
  507--515, 2013.

\bibitem[Schaul et~al.(2015)Schaul, Quan, Antonoglou, and
  Silver]{schaul2015prioritized}
Tom Schaul, John Quan, Ioannis Antonoglou, and David Silver.
\newblock Prioritized experience replay.
\newblock \emph{arXiv preprint arXiv:1511.05952}, 2015.

\bibitem[Schulman et~al.(2017)Schulman, Wolski, Dhariwal, Radford, and
  Klimov]{schulman2017proximal}
John Schulman, Filip Wolski, Prafulla Dhariwal, Alec Radford, and Oleg Klimov.
\newblock Proximal policy optimization algorithms.
\newblock \emph{arXiv preprint arXiv:1707.06347}, 2017.

\bibitem[Sutton and Barto(1998)]{sutton1998reinforcement}
R.~S. Sutton and A.~G. Barto.
\newblock \emph{Reinforcement learning: An introduction}, volume~1.
\newblock MIT press Cambridge, 1998.

\bibitem[Vapnik(2006)]{vapnik2006estimation}
Vladimir Vapnik.
\newblock \emph{Estimation of dependences based on empirical data}.
\newblock Springer Science \& Business Media, 2006.

\bibitem[Vitter(1985)]{vitter1985random}
Jeffrey~S Vitter.
\newblock Random sampling with a reservoir.
\newblock \emph{ACM Transactions on Mathematical Software (TOMS)}, 11\penalty0
  (1):\penalty0 37--57, 1985.

\bibitem[Wang and Ross(2019)]{wang2019boosting}
Che Wang and Keith Ross.
\newblock Boosting soft actor-critic: Emphasizing recent experience without
  forgetting the past.
\newblock \emph{arXiv preprint arXiv:1906.04009}, 2019.

\bibitem[Wixted and Ebbesen(1991)]{wixted1991form}
John~T Wixted and Ebbe~B Ebbesen.
\newblock On the form of forgetting.
\newblock \emph{Psychological science}, 2\penalty0 (6):\penalty0 409--415,
  1991.

\bibitem[Zenke et~al.(2017)Zenke, Poole, and Ganguli]{zenke2017continual}
F.~Zenke, B.~Poole, and S.~Ganguli.
\newblock Continual learning through synaptic intelligence.
\newblock In \emph{International Conference on Machine Learning}, pages
  3987--3995, 2017.

\bibitem[Zhang and Sutton(2017)]{zhang2017deeper}
S.~Zhang and R.S. Sutton.
\newblock A deeper look at experience replay.
\newblock In \emph{Deep Reinforcement Learning Symposium NIPS 2017}, 2017.

\end{thebibliography}
\bibliographystyle{plainnat}

\appendix
\renewcommand\thefigure{\thesection.\arabic{figure}}    
\setcounter{figure}{0}  
\section{Appendix}\label{app1}

\subsection{Linear Gravity Individual Evaluations} \label{app:additional}

\begin{figure}[h]
\centering
\begin{subfigure}{0.32\textwidth}
\includegraphics[width=\textwidth]{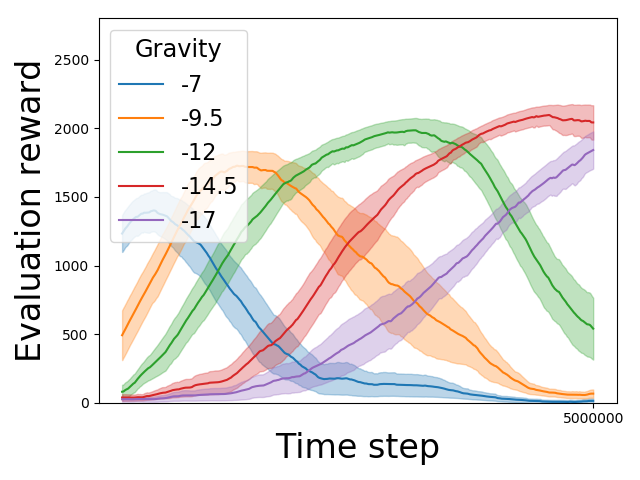}
\caption{FIFO}
\end{subfigure}
\begin{subfigure}{0.32\textwidth}
\includegraphics[width=\textwidth]{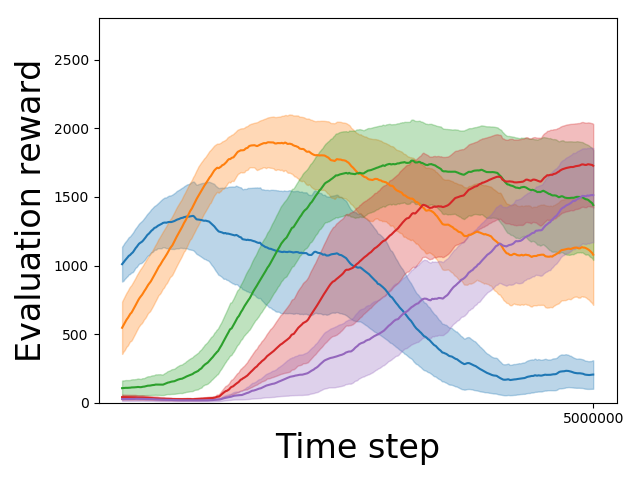}
\caption{Reservoir}
\end{subfigure}
\begin{subfigure}{0.32\textwidth}
\includegraphics[width=\textwidth]{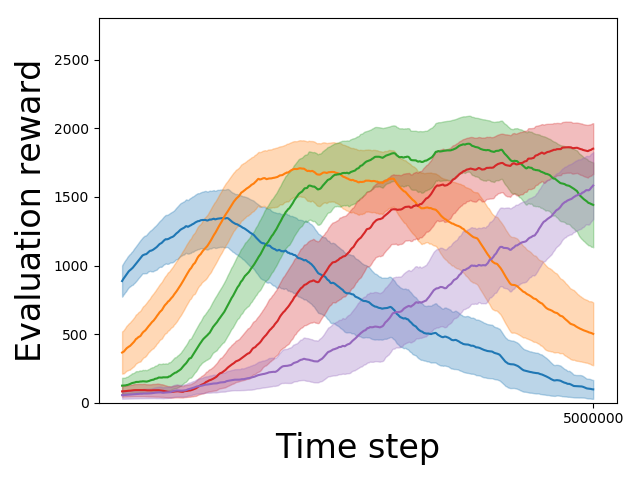}
\caption{Half Reservoir Half FIFO}
\end{subfigure}
\begin{subfigure}{0.32\textwidth}
\includegraphics[width=\textwidth]{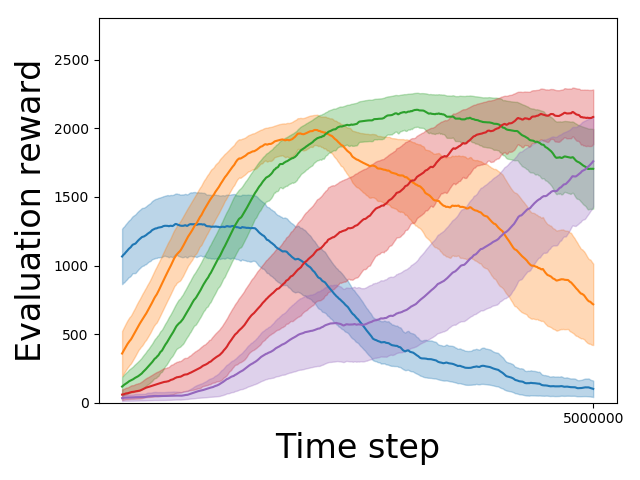}
\caption{MTR}
\end{subfigure}
\begin{subfigure}{0.32\textwidth}
\includegraphics[width=\textwidth]{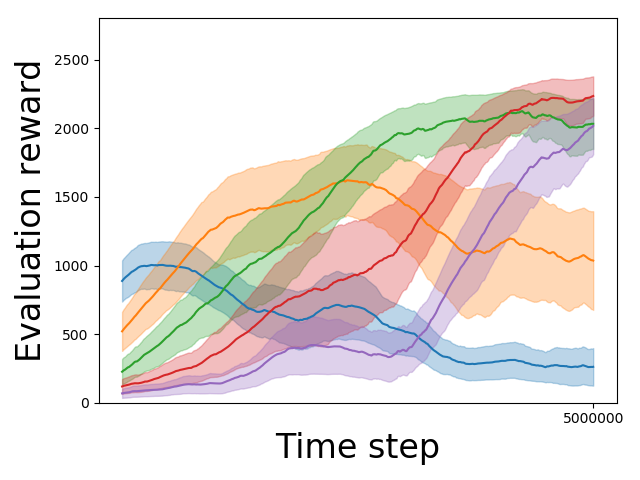}
\caption{MTR with IRM}
\end{subfigure}
\caption{Individual Evaluation rewards for linearly increasing gravity HalfCheetah. Mean and standard error bars over three runs.}
\label{fig:individual_eval_results}
\end{figure}

In Figure \ref{fig:individual_eval_results}(a), we can see that the cycles of learning and forgetting are quite clear with th FIFO agent. In all other agents, where older experiences were maintained for longer in the buffer, the forgetting process is slower. This does not seem to be qualitatively different for the MTR-IRM agent - it just seems to be able to reach a good balance between achieving a high performance in the various settings, while forgetting slowly. In particular, it is hard to identify whether there has been much \textit{forward transfer} to gravity settings that have yet to be trained on, which one might hope for by learning an invariant policy: at the beginning of training, the extra IRM constraints seem to inhibit the progress on all settings (as compared to the standard IRM agent), but in the latter stages the performance on a number of the later settings improves drastically. 

\subsection{Multi-task (Random gravity) Experiments} \label{app:multitask}

\begin{figure}[h]
\centering
\begin{subfigure}{0.49\textwidth}
\includegraphics[width=\textwidth]{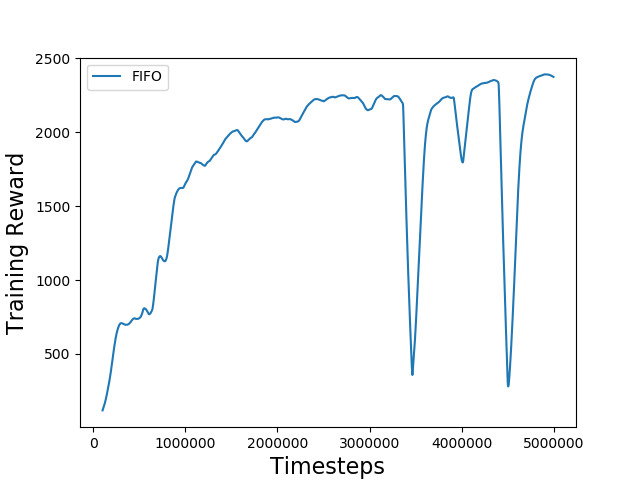}
\caption{Training performance}
\end{subfigure}
\begin{subfigure}{0.49\textwidth}
\includegraphics[width=\textwidth]{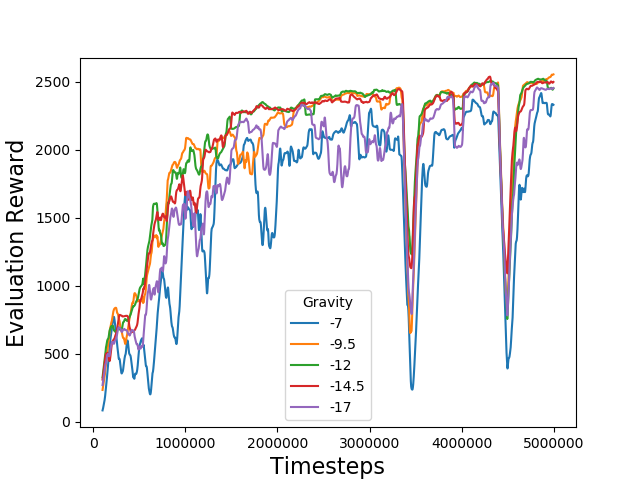}
\caption{Evaluation performance}
\end{subfigure}
\caption{Multitask setting: (a) training performance and (b) evaluation performance with uniformly random gravity between $-7$ and $-17m/s^2$ with a FIFO buffer. This experiment shows that the agent has the capacity to represent good policies for all evaluation settings if trained in a non-sequential setting.}
\label{fig:mtr_multitask}
\end{figure}

\subsection{Power Law Forgetting} \label{app:powerlaw}

Several studies have shown that memory performance in humans declines with a \textit{power law function} of time \citep{wixted1991form, rubin1996one}; in other words, the accuracy on a memory task at time $t$ is given by $y=a t^{-b}$ for some $a,b \in \mathbb{R}^{+}$ \citep{kahana2017note}. Here we provide a mathematical intuition for how the MTR buffer approximates a power law forgetting function of the form $\frac{1}{t}$, without giving a formal proof. If we assume the cascade is full, then the probability of an experience being pushed into the $k^{th}$ sub-buffer is ${\beta_{mtr}}^{k-1}$, since, for this to happen, one must be pushed from the $1^{st}$ to the $2^{nd}$ with probability $\beta_{mtr}$, and another from the $2^{nd}$ to the $3^{rd}$ with the same probability, and so on. So, in expectation, $\frac{N}{n_b}\cdot \frac{1}{{\beta_{mtr}}^{k-1}}$ new experiences must be added to the database for an experience to move from the beginning to the end of the $k^{th}$ sub-buffer. Thus, if an experience reaches the end of the $k^{th}$ buffer, then the expected number of time steps that have passed since that experience was added to the first buffer is given by:
\begin{align}
\hat{t}_k = \mathbb{E}[t | \text{end of $k^{th}$ buffer}] &= \sum_{i=1}^{k} \frac{N}{n_b} \cdot \frac{1}{{\beta_{mtr}}^{i-1}} \\
&= \frac{N}{n_b} \cdot \frac{\beta_{mtr}}{1-\beta_{mtr}} \cdot \left(\frac{1}{{\beta_{mtr}}^k}-1 \right) \label{eq:mtr_tk}
\end{align}
If we approximate the distribution $\mathbb{P}(t | \text{end of $k^{th}$ buffer})$ with a delta function at its mean, $\hat{t}_k$, and we note that the \textit{probability} of an experience making it into $(k+1)^{th}$ buffer at all is ${\beta_{mtr}}^k$, then, by rearranging Equation \ref{eq:mtr_tk}, we can say that the probability of an experience lasting more than $\hat{t}_k$ time steps in the database is given by:
\begin{align}
\mathbb{P}(\text{experience lifetime} > \hat{t}_k) = \frac{1}{\hat{t}_k \left(\frac{n_b}{N} \cdot \frac{1-\beta_{mtr}}{\beta_{mtr}}\right) + 1}
\end{align}
In other words, the probability of an experience having been retained after $t$ time steps is roughly proportional to $\frac{1}{t}$.

The expected number of experiences requires to fill up the MTR cascade (such that the size of the overflow buffer goes to zero) is calculated as follows: 
\begin{equation}
\sum_{i=1}^{n_b} \frac{N}{n_b} \cdot \frac{1}{{\beta_{mtr}}^{i-1}}
\end{equation}
which for $N=1e6$, $n_b=20$ and $\beta_{mtr}=0.85$, evaluates to 7 million experiences. 

\subsection{Experimental Details}\label{app:details}

\begin{figure}[h]
\centering
\begin{subfigure}{0.3 \textwidth}
\includegraphics[width=\textwidth]{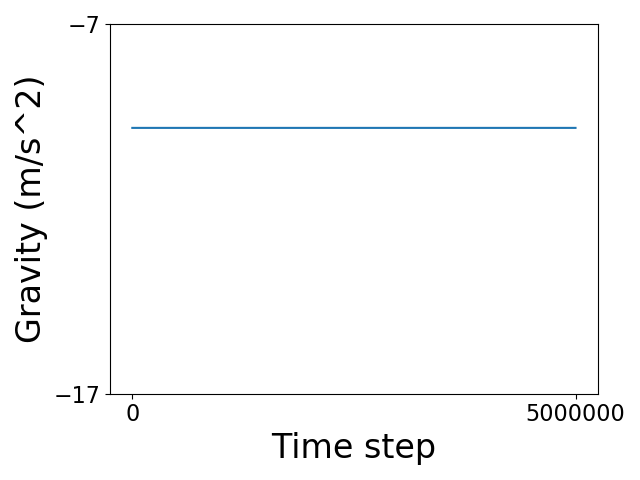}
\caption{Fixed gravity}
\end{subfigure}
\begin{subfigure}{0.3 \textwidth}
\includegraphics[width=\textwidth]{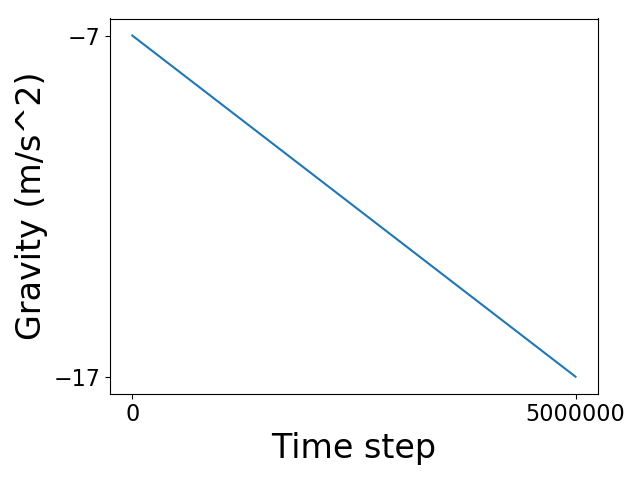}
\caption{Linearly increasing gravity}
\end{subfigure}
\begin{subfigure}{0.3 \textwidth}
\includegraphics[width=\textwidth]{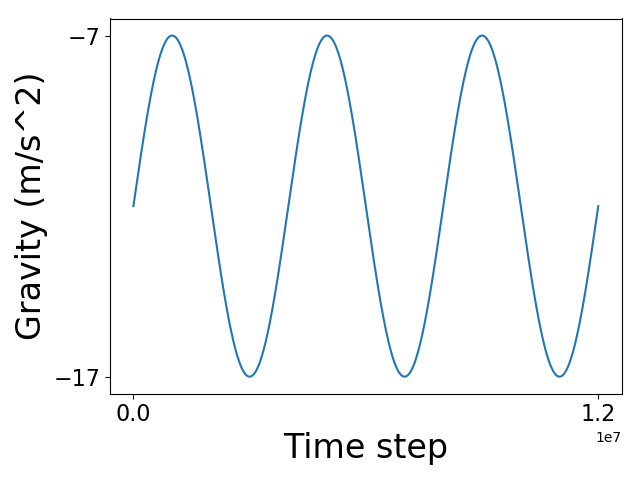}
\caption{Fluctuating gravity}
\end{subfigure}
\caption{Gravity settings over the course of the simulation in each of the three set-ups.} 
\label{fig:gravity}
\end{figure}

\paragraph{Gravity settings and baselines} The gravity changes in each of the different settings are shown in Figure \ref{fig:gravity}. The fixed and linear gravity experiments were run for 5 million time steps, but the fluctuating gravity was run for 12 million steps, with 3 full cycles of 4 million steps. The value of the gravity setting was appended to the state vector of the agent so that there was no ambiguity about what environment the agent was in at each time step. 

The MTR and MTR-IRM methods were compared with FIFO, reservoir and half-reservoir-half-FIFO baselines. In the last baseline, new experiences are pushed either into a FIFO buffer or a reservoir buffer (both of equal size) with equal probability, and sampled from . The maximum size of each database used is 1 million experiences and was chosen such that, in every experimental setting, the agent is unable to store the entire history of its experiences.

\paragraph{Hyperparameters} Below is a table of the relevant hyperparameters used in our experiments.

\begin{table*}[h]
\caption{Hyperparameters}
\label{sample-table}
\begin{center}
\begin{small}
\begin{sc}
\begin{tabular}{lcccr}
\toprule
Parameter & Value \\
\midrule

\# hidden layers (all networks) & 2 \\
\# units per hidden layer & 256 \\
Learning rate & 0.0003 \\
Optimiser & Adam \\
Adam $\beta_1$ & 0.9 \\
Adam $\beta_2$ & 0.999 \\
Replay Database size (all buffers) & 1e6 \\
\# MTR sub-buffers $n_b$ & 20 \\
$\beta_{mtr}$ & 0.85 \\
Hidden neuron type & ReLU \\
Target network $\tau$ & 0.005 \\
Target update frequency / time steps & 1 \\
Batch size & 256 \\
\# Training time steps & 5e6 (fixed), 5e6 (linear), 1.2e7 (fluctuating) \\
Training frequency / time steps & 1 \\
Gravity adjustment frequency / time steps & 1000 \\
Evaluation frequency / episodes & 100 \\
\# Episodes per evaluation & 1 \\
IRM policy coefficient & 0.1 \\
\bottomrule
\end{tabular}
\end{sc}
\end{small}
\end{center}
\label{table:params}
\end{table*}

\end{document}